# Research on Clustering Performance of Sparse Subspace Clustering

Wen-Jin Fu, Xiao-Jun Wu*, He-Feng Yin, Wen-Bo Hu

School of IoT Engineering, Jiangnan University, Wuxi 214122, China

**Abstract-**Recently, sparse subspace clustering has been a valid tool to deal with high-dimensional data. There are two essential steps in the framework of sparse subspace clustering. One is solving the coefficient matrix of data, and the other is constructing the affinity matrix from the coefficient matrix, which is applied to the spectral clustering. This paper investigates the factors which affect clustering performance from both clustering accuracy and stability of the approaches based on existing algorithms. We select four methods to solve the coefficient matrix and use four different ways to construct a similarity matrix for each coefficient matrix. Then we compare the clustering performance of different combinations on three datasets. The experimental results indicate that both the coefficient matrix and affinity matrix have a huge influence on clustering performance and how to develop a stable and valid algorithm still needs to be studied.

## 1. Introduction

The research on traditional clustering theory and methods has been relatively mature and achieved superb clustering results in many practical applications. In fact, representation of the high-dimensional unstructured image and video data has been a hot topic in the fields of computer vision, signal processing, and pattern recognition. In many application fields, high-dimensional data in the same class or directory can be well represented by low-dimensional subspaces. To reduce the dimensionality of original images, many subspace learning are presented [1-3]. In recent years, sparse representation has received widespread attention due to its wide application in signal processing [4-5].

Sparse subspace clustering (SSC) [6] uses the original dataset as a dictionary to obtain a sparse representation of data and constructs an affinity matrix using the sparse coefficient matrix to get the clusters. Generally, SSC is reduced to the following minimization problem:

$$\min_{Z} \alpha ||X - A(X)Z||_l + \Omega(X, Z)$$
$$\text{s.t. } Z \in \mathcal{C} \tag{1}$$

where $X \in \mathbb{R}^{d \times n}$ is data matrix, each column represents a data point, $A(X)$ represents a set of d-dimensional data vectors which is used as a dictionary, $||\cdot||_l$ is a proper norm, $\Omega(X, Z)$ and $\mathcal{C}$ are the regularization term and constraint set on $Z$, respectively, and $\alpha > 0$ is a trade-off parameter. At present, there are multiple norms to select for $||\cdot||_l$ and $\Omega(X, Z)$.

The current research on such algorithms is mainly focused on three aspects: regularization item design, residual item design and fast algorithm design. For regularization item design, SSC and low-rank representation (LRR) [7] use sparse and low-rank regularization to ensure coefficient matrices have block diagonal property. The coefficient matrix solved by SSC [6] and LRR has the sparse property and low-rank property, respectively. Least squares regression (LSR) uses Frobenius norm as the regularization item. Although LSR [8] has no sparse property like SSC, it satisfies the conditions of enhanced block diagonal (EBD) which can ensure the coefficient matrix has a block diagonal structure. Meanwhile, it has an analytical solution which can ensure the speed of the solution is faster than SSC and LRR. There are numerous improved methods according to the disadvantages of SSC, LRR and LSR. For example, Lu et al. [9] used Trace Lasso as the regularization term to solve the coefficient matrix. Xu et al. [10] proposed reweighted sparse subspace clustering (RSSC) which replaces $\ell_1$ norm with the reweighted $\ell_1$ norm in SSC. Tang et al. [11] combined the advantages of SSC and LSR by introducing $\ell_1$ norm and $\ell_2$ norm into the same objective function to propose dense block and sparse

*Corresponding author: xiaojun_wu_jnu@163.com

representation (DBSR). These algorithms focus on the design of the regularization item.

The residual item is used to minimize the error which can characterize noise and outlying entries. Currently, we use Frobenius norm to represent Gaussian noise and $\ell_{2,1}$ norm to represent singular samples. In order to deal with non-Gaussian noise and impulse noise, Lu et al. [12] introduced correntropy induced L2 (CIL2) method which uses correntropy induced metric to approximate error. The experimental results indicate that CIL2 is robust to non-Gaussian noise and impulse noise. In addition, we can obtain a clean data matrix using some existing methods such as robust principal component analysis (RPCA) [13] and principal component analysis (PCA) [14] for the data with noise firstly and then use SSC or LSR to solve the coefficient matrix. It has proved that a clean dictionary can improve the robustness of clustering in LRSC [15]. SSC and LRR have no analytical solution, so alternating direction method of multipliers (ADMM) [16] algorithm is used to solve the coefficient matrix. However, the ADMM method needs a great number of iterations until the algorithm converges and the time complexity is high for LRR method at each iteration. So for fast algorithm design, linear ADMM [17] is always used to speed up the solving of the $\ell_1$ or nuclear minimization problems. Patel et al. [18] used SSC to cluster the data after the data is reduced into low-dimensional space. This method is named as latent space SSC. It decreases the running time of the algorithm by reducing the dimension of the data.

A lot of experiments demonstrate the effectiveness of the existing improved algorithms for sparse subspace clustering. But there are few literatures study the influence factors of sparse subspace clustering and which is essential, coefficient matrix or affinity matrix? In our experiments, for the same coefficient matrix, when we select different methods to construct an affinity matrix, the accuracy of clustering has twenty percent gap. This paper studies the influence of the coefficient matrix and affinity matrix for clustering performance. We select four methods to solve the coefficient matrix of data including SSC, LSR, smooth representation clustering (SMR) [19] and low-rank representation with symmetric constraint (LRRSC) [20] respectively. SSC and LSR are the classic subspace clustering approaches. SMR can be seen as the improvement on LSR, which has a grouping effect. LRRSC is the improvement on LRR, which can guarantee the symmetry of the coefficient matrix and reflect the subspace structure of data. The four methods of constructing affinity matrix are different: symmetrical method (SM) is used in SSC, sparse symmetrical method (SSM) is used in thresholding ridge regression (TRR), SVD method (SVDM) is used in LRRSC, and inner product method (IPM) is used in SMR.

The remainder of this paper is organized as follows. Section 2 briefly introduces the methods of solving the coefficient matrix and constructing an affinity matrix. Section 3 discusses the factors which influence the clustering performance. Finally, Section 4 concludes this paper.

**2. Related algorithms and principles**

Section 2.1 introduces SSC, LSR, SMR and LRRSC. Section 2.2 presents the methods of constructing an affinity matrix, which includes SM, SSM, SVDM and IPM.

**2.1 The method of solving coefficient matrix**

SSC uses $\ell_1$ norm as regularization term to solve the coefficient matrix. The objective function is given by

$$\min_{C} ||C||_1 + \lambda_e ||E||_1 + \frac{\lambda_z}{2}||Z||_F^2 \\ \text{s.t.} \ \ X = XC + E + Z, \text{diag}(C) = 0, \qquad (2)$$

where $\ell_1$-norm promotes sparsity of the columns of $C$ and $E$, while the Frobenius norm promotes having small entries in the columns of $Z$. The two parameters $\lambda_e$ and $\lambda_z$ balance the three terms in the objective function. $\mathrm{diag}(C)$ ensures the diagonal entries are 0. The coefficient matrix solved by SSC is sparse and non-zero elements represent the relationship between data points.

LSR use Frobenius norm as regularization term, and it has a grouping effect. It has proved that in LSR the coefficients of the data points in the same subspace are larger than those in the other subspaces. It has low time complexity due to the analytic solution. The objective function is

$$\min_C ||X - XC||_F^2 + \lambda ||C||_F^2$$
$$\text{s.t. } \mathrm{diag}(Z) = 0 \tag{3}$$

where $\lambda > 0$ is a parameter used to balance the effects of the two items. $||C||_F^2$ denotes the Frobenius norm of $C$. $||X - XC||_F^2$ denotes the self-reconstruction error.

SMR uses $\mathrm{tr}(C\hat{L}C)$ as a regularization item to solve the coefficient matrix. Like LSR, SMR also has a grouping effect, and the coefficient matrix solved by SMR can protrude the local characteristics of the data. Compared with SSC, SMR can make the relationship within the class denser. The objective function of SMR is

$$\min_C \lambda ||X - XC||_F^2 + \mathrm{tr}(C\widetilde{L}C^T), \tag{4}$$

where $\hat{L} = L + \epsilon I$, $L = D - W$ is a Laplacian matrix which is used to protrude the local characteristics. $W$ is the similarity matrix of dataset $X$. $D$ is the degree matrix, where $D_{ii} = \sum_{j=1}^n w_{ij}$.

LRRSC introduces the symmetric constraint into the objective function of LRR and makes the coefficient matrix reflect the subspace structure. Like LRR, this method needs a mass of iterations and has high time complexity. The objective function of LRRSC is

$$\min_{C,E} ||C||_* + \lambda ||E||_{2,1}$$
$$\text{s.t.} X = AC + E, C = C^T, \tag{5}$$

where $A$ is a low-rank matrix derived from the given set of data. $||\cdot||_*$ denotes the nuclear norm. $||E||_{2,1}$ is used to characterize the error term. The symmetric constraint ensures the symmetry of $C$.

**2.2 The method of constructing affinity matrix**

We can use many methods to construct the affinity matrix. The most classic method is proposed by Elhamifar and Vidal in the SSC model. In order to build the symmetric affinity matrix, SSC defines an affinity matrix as

$$W = \frac{|C| + |C|^T}{2}, \tag{6}$$

where $C$ is the coefficient matrix solved by SSC. This method is simple and has been applied to SSC and LSR. Because of the symmetry of the coefficient matrix, we named it as a symmetrical method (SM).

The coefficient matrix solved by LSR has no property of sparsity, so the affinity matrix formed by (6) is not sparse which is detrimental to spectral clustering. Peng et al. [21] proposed a new method to construct the affinity matrix which obtains a sparse, and symmetric affinity matrix for spectral clustering by

$$W = \frac{|\hat{C}| + |\hat{C}|^T}{2}, \tag{7}$$

where $\hat{C}$ is generated by keeping the k largest entries in each column of $C$ and the others are set to zeros. By this way, we can get a sparse and symmetric affinity matrix and we name it as the sparse symmetrical method (SSM).

Due to the presence of noise, formula (5) cannot reflect the relationship between data correctly.

Chen et al. [20, 22] consider $C$ with the skinny SVD $U^* \sum^* (V^*)^T$ and then use $U^*(\sum^*)^{\frac{1}{2}}$ or $(\sum^*)^{\frac{1}{2}}(V^*)^T$ to construct the affinity matrix. The method of affinity matrix is given by

$$W_{ij} = \left(\frac{m_i^T m_j}{||m_i||_2 ||m_j||_2}\right)^{2\alpha} \text{or} W_{ij} = \left(\frac{n_i^T n_j}{||n_i||_2 ||n_j||_2}\right)^{2\alpha}, \tag{8}$$

where $m_i$ and $m_j$ are the ith and jth row of $M = U(\sum)^{1/2}$, respectively. $n_i$ and $n_j$ are the ith and jth column of $N = (\sum)^{1/2} V^T$, respectively. $\alpha$ is a parameter to adjust the element of $W$. We name this method as the SVD method (SVDM).

In order to exploit the merit of grouping effect, Hu et al. defined a new affinity matrix. The method of affinity matrix is given by

$$W_{ij} = \left(\left|\frac{c_i^T c_j}{||x_i||_2 ||x_{j2}||}\right|^\alpha\right), \tag{9}$$

where $c_i^T c_j$ represents the inner product of vectors. Under the assumption that the linear subspaces are independent, we can know that $c_i^T c_j$ is zero if $x_i$ and $x_j$ come from different subspaces. We name this method as inner product method (IPM).

**3. Factors affecting clustering performance**

This section studies how the coefficient matrix and affinity matrix affect clustering performance from two aspects. Section 3.1 introduces the datasets used in the experiments. Section 3.2 gives the parameters of compared algorithms. Section 3.3 investigates the influence from the viewpoint of clustering accuracy and Section 3.4 investigates the influence using the metric of standard deviation from the perspective of stability. All the experimental data are given in Section 3.5. All the algorithms are performed on a personal computer with Intel Core i3-3240 CPU and 8GB RAM which are implemented with Matlab 2013a.

**3.1 Datasets and experimental settings**

To investigate the influence of coefficient matrix and affinity matrix on clustering performance, we conduct different experiments on three popular benchmark databases, i.e., the Extended Yale B [23], AR [24] and USPS [25] respectively. The description of the three databases is summarized a follows.

- Extended Yale B contains 2414 frontal images of 38 individuals, with images of each individual lying in a low-dimensional subspace. There are 59-64 images available for each individual. In order to reduce the computational time and memory requirements of the algorithm, we use a normalized face image with size $48 \times 42$ pixels in the experiments. We select face images of the first 10 subject for the experiments, and the data are projected into a 10 * 6-dimensional subspace by PCA.
- AR contains over 4000 images of 70 male subjects and 56 female subjects. Each subject has 26 images which are taken in two separate sessions. These images suffer different facial variations, including various facial expressions (neutral, smile, anger and scream), illumination variations (left light on, right light on and side lights on), and occlusion by sunglasses or scarf. We select face images of the first 20 subjects for the experiments, and the data are projected into 20 * 6-dimensional subspace by PCA.
- USPS is a handwritten digit dataset with 9298 images, with each image having $16 \times 16$ pixels. We select the first 100 images of each digit for experiments.

We use the mean and standard deviation of clustering accuracy to evaluate the performance of clustering. All experiments are repeated 20 times to calculate the mean clustering accuracy and standard deviation.

**3.2 Parameters setting**

For the Extended Yale B dataset: In LSR, we set $\lambda = 0.01$. When using SSM to construct the affinity matrix, we set k=5. For SVDM, we use $\alpha = 3$. For IPM, we use $\alpha = 6$. In SMR, we set $\lambda = 2^{15}$. When using SSM to construct the affinity, we set k=5. For SVDM, we use $\alpha = 5$. For IPM, we use $\alpha = 5$. In LRRSC, we set $\lambda = 0.2$. For SSM, we set k=7. For SVDM, we use $\alpha = 4$. For IPM, we use $\alpha = 3$. In SSC, we set $\lambda = 20$. For SSM, we set k=5. For SVDM, we use $\alpha = 2$. For IPM, we use $\alpha = 3$.

For the AR dataset: In LSR, we set $\lambda = 0.01$. For SSM, we set k=5. For SVDM, we use $\alpha = 1$. For IPM, we use $\alpha = 1$. In SMR, we set $\lambda = 2^{20}$. For SSM, we set k=5. For SVDM, we use $\alpha = 1$. For IPM, we use $\alpha = 5$. In LRRSC, we set $\lambda = 2$. For SSM, we set k=5. For SVDM, we use $\alpha = 1$. For IPM, we use $\alpha = 1$. In SSC, we set $\lambda = 20$. For SSM, we set k=8. For SVDM, we use $\alpha = 1/8$. For IPM, we use $\alpha = 1$.

For the USPS dataset: In LSR, we set $\lambda = 5$. For SSM, we set k=7. For SVDM, we use $\alpha = 3$. For IPM, we use $\alpha = 1$. In SMR, we set $\lambda = 2^{-16}$. For SSM, we set k=5. For SVDM, we use $\alpha = 3$. For IPM, we use $\alpha = 1$. In LRRSC, we set $\lambda = 0.001$. For SSM, we set k=7. For SVDM, we use $\alpha = 4$. For IPM, we use $\alpha = 2$. In SSC, we set $\lambda = 10$. For SSM, we set k=8. For SVDM, we use $\alpha = 1$. For IPM, we use $\alpha = 4$.

### 3.3 Clustering accuracy

Figure 1 depicts the influence of coefficient matrix and affinity matrix over clustering accuracy on the Extended Yale B, AR and USPS datasets.

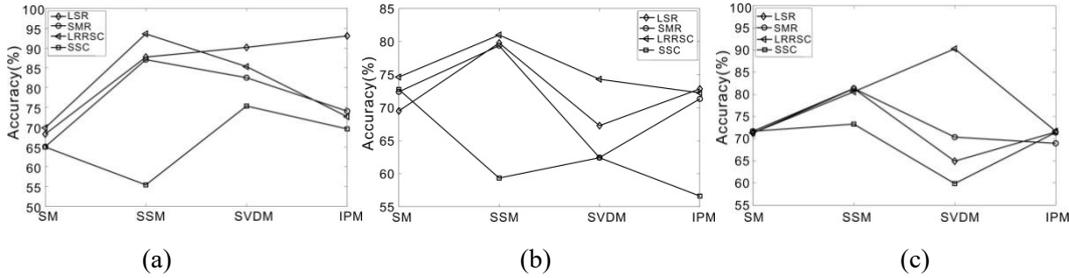

(a)　　　　　　　　　　　(b)　　　　　　　　　　　(c)

Figure 1: The influence of different combinations on clustering accuracy on different datasets. (a) The influence of coefficient matrix and affinity matrix on clustering accuracy on the Extended Yale B dataset. (b) The influence of coefficient matrix and affinity matrix on clustering accuracy on the AR dataset. (c) The influence of coefficient matrix and affinity matrix on clustering accuracy on the USPS dataset.

It can be seen from Figure 1(a) that when we use different methods to construct the coefficient matrix for the same affinity matrix, the clustering accuracy exhibits a huge difference on the Extended Yale B dataset. For example, when using LRRSC and SSM, we can obtain the best clustering accuracy of 93.59%. Nevertheless, if we select SSC and SSM to cluster the data, the accuracy drops to 55.41% which is the worst. The difference between the two combinations is 38.18% approximately. On the Extended Yale B dataset, when using the SVDM method to construct the affinity matrix, the clustering accuracy fluctuates less. The accuracy of SSC and SVDM is 75.35%, while LSR and SVDM is 90.17%. The difference is 14.82%. We can also see that on the Extended Yale B dataset, regardless of which way to build a similarity matrix, SSC has the poor clustering accuracy. For the same coefficient matrix, when using different methods of building affinity matrix, the accuracy of clustering fluctuates greatly. For SSC, the accuracy of SSM and SVDM is 55.41% and 75.35%, respectively. The difference between SSM and SVDM is 19.94%. For LSR, the difference is 24.81%. For SMR, the difference is 22%. For LRRSC, the difference is 23.75%.

From Figure 1(b), we can see that on the AR dataset, it has a similar phenomenon as on the Extended Yale B dataset. Just as on the Extended Yale B dataset, LRRSC and SSM can yield the best clustering accuracy of 81%. For the same coefficient matrix, when using different methods to build the affinity matrix, the clustering accuracy fluctuates significantly. For SSC, using the SM method can get the best clustering accuracy of 72.76%, and the IPM is the worst at 56.62%. The difference is 16.14%. For LSR, SSM obtains the best result of 79.79%, while SVDM gets the worst result of 67.28%. The difference is 12.51%. For SMR, it is the same as LSR that SSM gets the best accuracy of 79.38%, while SVDM gets the worst result of 62.44%. The difference is 16.94%. For LRRSC, the difference is 8.76% which is better than that of the others. When fixing the method of building an affinity matrix, we can know the influence of the coefficient matrix over accuracy on the AR dataset. For SM, different coefficient matrices have similar clustering accuracy. The maximum difference in accuracy is only 3.27% which is the least compared with the other methods. For the SSM method, LSR, SMR and LRRSC have similar clustering accuracy, but SSC has a huge gap in clustering accuracy compared with them. For SVDM and IPM, all the combinations have poor results.

From Figure 1(c), we can observe that it has a different phenomenon on the USPS dataset compared with Extended Yale B and AR datasets. The combination of LRRSC and SSM can get the best clustering accuracy on the Extended Yale B and AR datasets. However, on USPS dataset, the combination of LRRSC and SVDM gets the best result of 90.3%. It is interesting that when LSR, SMR and SSC are combined with SVDM, respectively, the clustering accuracy generated by these combinations is poor. On the USPS dataset, it seems that different methods of building affinity matrix except SVDM has less influence on the clustering accuracy. For example, when choosing SM to build an affinity matrix, SMR can get the best clustering accuracy of 71.7%. In contract, LSR gains the worst clustering accuracy of 71.27%. The difference is only 0.43%. For SSM, the difference is 8.04%. For IPM, the difference is 2.55%. Like the other datasets, on the USPS dataset, when we fix the method of building the coefficient matrix, different methods of establishing affinity matrix have a huge influence on the clustering accuracy. LRRSC and SVDM can get the best accuracy 90.3%, while SSC and SVDM obtain the worst accuracy of 59.88%. The difference is 30.42%.

From what has been analyzed above, we can make the following observations:

(1) To solve different problems, after designing a method to solve the coefficient matrix, we should choose an appropriate method to build the affinity matrix to get the best clustering accuracy. For example, on the Extended Yale B dataset, using IPM to build the affinity matrix can get the best accuracy for LSR method. On the AR dataset, LSR and SSM can get the best clustering accuracy. In practice, it is hard for us to select a uniform and valid method to build an affinity matrix for the same coefficient matrix on different dataset. So can we design a uniform and valid algorithm framework of building the affinity matrix for the same coefficient matrix to reduce the impact of the affinity matrix on clustering accuracy on different datasets while obtaining the best clustering accuracy?

(2) For the same coefficient matrix, the results generated by different methods of constructing affinity matrix have a huge difference on the same dataset. For example, on the Extended Yale B database, SSM is the best method of building affinity matrix for LRRSC. LRRSC and SM get the worst clustering accuracy of 69.84%. It indicates that the research on the method of building an affinity matrix is insufficient. Thus, in future work, we should pay more attention to the affinity matrix in order to design a method to build the affinity matrix, which can reduce its effect on the clustering accuracy.

(3) Both the methods of solving the coefficient matrix and building the affinity matrix influence the clustering accuracy. (2) has suggested designing a valid method of building the affinity matrix to reduce the impact of affinity matrix on clustering accuracy. Now, can we design a new method of solving the coefficient matrix to reduce the effect of the coefficient matrix on the clustering accuracy?

## 3.4 Stability

Figure 2 shows the influence of coefficient matrix and affinity matrix over standard deviation on the Extended Yale B, AR and USPS datasets.

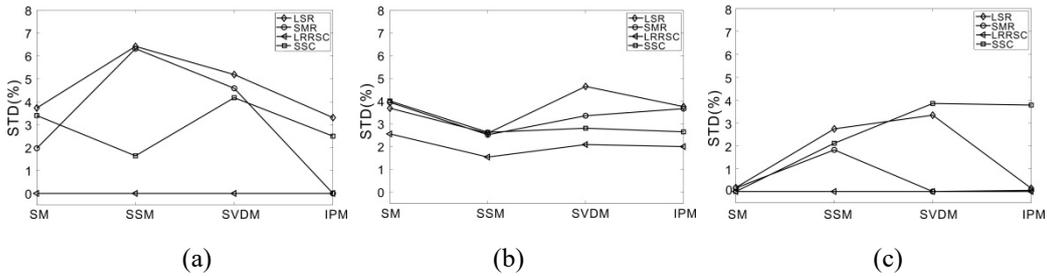

(a)　　　　　　　　　　　　(b)　　　　　　　　　　　　(c)

Figure 2: The influence of different combinations on standard deviation on different datasets. (a) The influence of coefficient matrix and affinity matrix on standard deviation on the Extended Yale B dataset. (b) The influence of coefficient matrix and affinity matrix on standard deviation on the AR dataset. (c) The influence of coefficient matrix and affinity matrix on standard deviation on the USPS dataset.

From Figure 2(a), we can see that different combinations have different influences on the stability of algorithms on the Extended Yale B dataset. Nevertheless, it is obvious that no matter which method of building the affinity matrix is used, the coefficient matrix solved by LRRSC has the least clustering standard deviation of 0%. We can see that the standard deviation solved by the combination of LRRSC and SM is 0%. LRRSC and SSM also achieve 0% standard deviation. LRRSC and SVDM is 0% and LRRSC with IPM is 0%. But when we use other methods to solve the coefficient matrix, the similarity matrix has a great influence on the stability of clustering and the standard deviation achieved by all other combinations is relatively large. Like SMR, using SM to build the affinity matrix can get 1.97% standard deviation. SSM is 6.33%. SVDM is 4.59%. IPM is 0%. The standard deviation changes greatly when using SMR to solve the coefficient matrix for different methods of building affinity matrix. No matter which method of building affinity matrix is used, LSR has the maximum standard deviation. The largest standard deviation is 6.33%.

From Figure 2(b), we can see that the fluctuation of standard deviation solved by different combination is relatively small on the AR dataset. When the method of solving the coefficient matrix is fixed, the stability of the algorithm is almost the same by using different methods to build the affinity matrix. Like LSR, the maximum and minimum standard deviation achieved by using different methods of building affinity matrix are 4.65% and 2.59%, respectively. So the method of building the affinity matrix has little effect on the stability of the algorithm when the coefficient matrix is fixed on the AR dataset. Meanwhile, like on the Extended Yale B dataset, there is a similar phenomenon on the AR dataset. For the LRRSC method of generating the coefficient matrix, regardless of which way to establish a similarity matrix, the standard deviations achieved by these combinations are small and stable. For example, the combination of LRRSC and SM achieves 2.56% standard deviation. LRRSC and SSM is 1.54%, LRRSC and SVDM is 2.09%. LRRSC and IPM is 2%. The maximum difference is only 1.02%.

From Figure 2(c), on the the USPS dataset, we can see that it has a similar phenomenon like on the Extended Yale B and AR datasets. For example, when LRRSC is used to solve the coefficient matrix, whichever method we choose to build the affinity matrix, the standard deviation of the algorithm is the same and at least 0%. We can also find that for other methods of generating the coefficient matrix, when the proper method of building affinity matrix is used, the standard deviation can also achieve 0% on the USPS dataset. Like SMR, when SVDM is used to construct the affinity matrix, the standard deviation of the algorithm is 0%. For SSC method, when SM is used to construct the affinity matrix, the standard deviation of the algorithm is 0%. When other methods are used to solve the coefficient matrix except for LRRSC, the standard deviations solved by different affinity matrix fluctuate greatly. For LSR, the standard deviation of SM is 0.17%, SSM is 2.74%, SVDM is 3.34% and IPM is 0.14%.

From the above analysis, we can make the following observations:

(1) Both the coefficient matrix and affinity matrix have an impact on the stability of the algorithm. But if we design an advanced method of solving the coefficient matrix-like LRRSC, the method of building an affinity matrix has no effect on the stability of the algorithm. So we still need to work in the direction of searching a better algorithm of solving the coefficient matrix which can improve the stability of the algorithm like LRRSC.

(2) From the above discussion, it seems that the method of building an affinity matrix is not as important as the method of solving the coefficient matrix for the stability of the algorithm. But we think that designing a better method of building the affinity matrix still makes sense for improving the stability of the algorithm for some methods of solving the coefficient matrix. For example, on the Extended Yale B dataset, the standard deviations solved by SMR with SM, SSM, SVDM are 1.97%, 6.33% and 4.59%, respectively, but if IPM is used to construct the affinity matrix, the standard deviation will be 0%. So when given the method of solving the coefficient matrix, choosing different methods to build affinity matrix can adjust the stability of the algorithm.

From the discussions in Section 3.3, we know that both coefficient matrix and affinity matrix have an effect on the clustering accuracy. How to combine the two items should be further studied. Based on the current methods of solving the coefficient matrix, we hope to find a unified framework to construct the affinity matrix in order to reduce the influence of affinity matrix on the clustering accuracy. The method of solving the coefficient matrix still needs to be further studied.

From the analysis in Section 3.4, we believe that the method of solving the coefficient matrix is more important than the method of building the affinity matrix on standard deviation of clustering. When a good method of solving the coefficient matrix is designed, the method of building an affinity matrix has no effect on the stability of clustering. But designing an affinity matrix to improve the stability of clustering is still significant when given a poor method of solving the coefficient matrix.

**3.5 Experimental data**

Tables 1-3 give the experimental data of all combinations on the above three datasets.

From Table 1, we can find some interesting results. For the SM method, LRRSC can get the best mean clustering accuracy of 69.84%. Meanwhile, LRRSC gains 0% standard deviation which is the least. But on the metric of max clustering accuracy, LSR gets the best clustering accuracy of 76.09%. Meanwhile, LRRSC is 69.84%. Then LRRSC gets the best result on the minimum clustering accuracy. When other methods of building affinity matrix are used, we can get a similar result. So we can know that when the method of building an affinity matrix is fixed, different coefficient matrices have a different effect on the clustering performance.

For LSR, when IPM is employed to build the affinity matrix, the best result can be derived on mean clustering accuracy and on clustering standard deviation. But SVDM gains the best result on max and min clustering accuracy. For LRRSC, SSM gets the best results on all metrics. On mean, max and min clustering accuracy, all the values are 93.59%. On clustering standard deviation, the result is 0%. When SM is used, the mean clustering accuracy is 69.84%. Compared with SSM, the gap between the two methods is 23.75%. It demonstrates that a good method of building affinity matrix can also improve the clustering performance dramatically.

From Tables 2-3, we can draw similar conclusions with Table 1.

Table 1. The results of different combinations on the Extended Yale B dataset

| Method | indicator | LSR | SMR | LRRSC | SSC |
|---|---|---|---|---|---|
| SM | Mean | 68.26 | 65.07 | 69.84 | 64.98 |
| | STD | 3.73 | 1.97 | **0** | 3.4 |
| | Max | 76.09 | 72.97 | 69.84 | 69.38 |
| | Min | 64.69 | 63.28 | 69.84 | 56.41 |
| SSM | Mean | 87.72 | 87.07 | **93.59** | 55.41 |
| | STD | 6.42 | 6.32 | **0** | 1.64 |
| | Max | 96.25 | 96.56 | 93.59 | 59.84 |
| | Min | 81.72 | 82.03 | 93.59 | 51.25 |
| SVDM | Mean | 90.17 | 82.52 | 85.31 | 75.35 |
| | STD | 5.19 | 4.59 | **0** | 4.18 |
| | Max | 96.88 | 86.72 | 85.31 | 79.38 |
| | Min | 85.47 | 76.25 | 85.31 | 65 |
| IPM | Mean | 93.07 | 74.063 | 72.66 | 69.56 |
| | STD | 3.31 | 4.59 | **0** | 2.5 |
| | Max | 95.16 | 74.06 | 72.66 | 72.97 |
| | Min | 85 | 76.25 | 72.66 | 64.84 |

Table 2. The results of different combinations on the AR dataset

| Method | indicator | LSR | SMR | LRRSC | SSC |
|---|---|---|---|---|---|
| SM | Mean | 69.49 | 72.40 | 74.64 | 72.76 |
| | STD | 3.69 | 3.95 | 2.56 | 4.01 |
| | Max | 76.54 | 81.15 | 79.23 | 79.81 |
| | Min | 63.08 | 66.92 | 70.58 | 65.77 |
| SSM | Mean | 79.79 | 79.38 | **81.00** | 59.35 |
| | STD | 2.59 | 2.52 | **1.54** | 2.63 |
| | Max | 83.65 | 82.69 | 85.00 | 63.65 |
| | Min | 74.04 | 73.27 | 77.89 | 55.77 |
| SVDM | Mean | 67.28 | 62.44 | 74.30 | 62.42 |

|  | STD | 4.65 | 3.36 | 2.09 | 2.81 |
|---|---|---|---|---|---|
|  | Max | 75.77 | 69.23 | 76.73 | 68.08 |
|  | Min | 58.85 | 54.42 | 70.19 | 58.08 |
| IPM | Mean | 72.84 | 71.33 | 72.24 | 56.62 |
|  | STD | 3.76 | 3.67 | 2.00 | 2.65 |
|  | Max | 78.65 | 69.23 | 77.89 | 61.73 |
|  | Min | 63.85 | 54.42 | 68.27 | 51.54 |

Table 3. The results of different combinations on the USPS dataset

| Method | indicator | LSR | SMR | LRRSC | SSC |
|---|---|---|---|---|---|
| SM | Mean | 71.27 | 71.69 | 71.30 | 71.70 |
|  | STD | 0.17 | 0.15 | 0 | 0 |
|  | Max | 71.50 | 72.20 | 71.30 | 71.70 |
|  | Min | 71.10 | 71.50 | 71.30 | 71.70 |
| SSM | Mean | 81.30 | 81.34 | 80.60 | 73.30 |
|  | STD | 2.74 | 1.82 | **0** | 2.11 |
|  | Max | 82.40 | 84.50 | 80.60 | 80.20 |
|  | Min | 73.10 | 76.20 | 80.60 | 72 |
| SVDM | Mean | 64.93 | 70.60 | **90.30** | 59.88 |
|  | STD | 3.34 | 0 | **0** | 3.85 |
|  | Max | 68.30 | 70.60 | 90.30 | 64.30 |
|  | Min | 60.30 | 70.60 | 90.30 | 55.00 |
| IPM | Mean | 71.51 | 68.96 | 71.70 | 71.41 |
|  | STD | 0.14 | 0.05 | **0** | 3.77 |
|  | Max | 71.70 | 69.00 | 71.70 | 79.60 |
|  | Min | 71.40 | 68.90 | 71.70 | 66.30 |

**4. Conclusion**

We have studied the impact of the coefficient matrix and affinity matrix on the clustering performance from two aspects, i.e., clustering accuracy and standard deviation. From what has been studied in Section 3, we can draw the following conclusions. (1) There is not enough research on the method of building an affinity matrix which is important for clustering accuracy as well as a coefficient matrix. (2) For the stability of the algorithm, the method of solving the coefficient matrix is more important than the method of constructing the affinity matrix. (3) How to combine the methods of solving the coefficient matrix and building affinity matrix is a worthwhile direction to be studied.